\documentclass{article}
\usepackage{arxiv}
\usepackage{amsfonts,booktabs,multirow,lineno,hyperref,amsmath}
\usepackage{blkarray, bigstrut,pifont}
\usepackage{booktabs}
\usepackage{amsthm,subfigure}
\usepackage{hyperref,booktabs}
\usepackage{algorithmic}
\usepackage{algorithm}
\usepackage[table,xcdraw]{xcolor}
\usepackage{pdflscape}

\usepackage[utf8]{inputenc} 
\usepackage[T1]{fontenc}    
\usepackage{hyperref}       
\usepackage{url}            
\usepackage{booktabs}       
\usepackage{amsfonts}       
\usepackage{nicefrac}       
\usepackage{microtype}      
\usepackage{lipsum}
\usepackage{graphicx}
\usepackage[shortlabels]{enumitem}
\usepackage{graphicx,amssymb,url,amsfonts,amsmath,mathrsfs,epsf,subfigure}
\usepackage{algorithm,booktabs,epstopdf,hyperref,pifont}
\usepackage{algorithmic}
\usepackage{threeparttable,lineno,multicol,comment}
\usepackage{rotating} 
\usepackage{amssymb}
\usepackage{amsfonts}
\usepackage{siunitx}
\usepackage{url}
\usepackage{color}
\usepackage{setspace}
\usepackage{diagbox}
\usepackage{multicol}
\usepackage{multirow}
\usepackage{float}
\usepackage{booktabs}

\usepackage{amsmath,amssymb,bm}
\usepackage{caption}
\usepackage{subfigure}
\usepackage{multirow}
\usepackage[english]{babel}
\usepackage{amsthm}

\title{Scalable Property Valuation Models via Graph-based Deep Learning}

\author{
Enrique Riveros\\
Universidad de los Andes, Chile \\
Facultad de Ingenier\'{i}a y Ciencias Aplicadas\\
\texttt{erriveros@miuandes.cl}
\And
Carla Vairetti \\
Universidad de los Andes, Chile \\
Facultad de Ingenier\'{i}a y Ciencias Aplicadas\\
\& Instituto Sistemas Complejos de Ingenier\'\i a (ISCI), Chile.\\
\texttt{cvairetti@uandes.cl}
\And
Christian Wegmann \\
ESE Business School, Universidad de los Andes, Chile \\
\texttt{cwegmann.ese@uandes.cl}
\And
Santiago Truffa\\
ESE Business School, Universidad de los Andes, Chile \\
\texttt{struffa.ese@uandes.cl}
\And
Sebasti\'an Maldonado\\
Department of Management Control and Information Systems \\
School of Economics and Business, University of Chile, Chile\\
\& Instituto Sistemas Complejos de Ingenier\'\i a (ISCI), Chile.\\
\texttt{sebastianm@fen.uchile.cl}
}

\begin{document}
\maketitle

\begin{abstract}
This paper aims to enrich the capabilities of existing deep learning-based automated valuation models through an efficient graph representation of peer dependencies, thus capturing intricate spatial relationships. In particular, we develop two novel graph neural network models that effectively identify sequences of neighboring houses with similar features, employing different message passing algorithms. The first strategy consider standard spatial graph convolutions, while the second one utilizes transformer graph convolutions. This approach confers scalability to the modeling process. The experimental evaluation is conducted using a proprietary dataset comprising approximately 200,000 houses located in Santiago, Chile. We show that employing tailored graph neural networks significantly improves the accuracy of house price prediction, especially when utilizing transformer convolutional message passing layers.
\end{abstract}
\footnote{This is a preprint of a work under submission and thus subject to change. Changes resulting from the publishing process, such as editing, corrections,structural formatting, and other quality control mechanisms may not be reflected in this version of the document.}

\section{Introduction}

The accurate evaluation of property prices plays an indispensable role in various economic facets of society. Serving as a foundational pillar of any economy, the real estate market drives significant property transactions, contributing substantially to a country's gross domestic product (GDP) \cite{sayce2009real}. 

Beyond its macroeconomic implications, on a personal financial scale, property acquisition stands as the most substantial investment for many citizens. Precise property valuations become paramount, ensuring equitable transactions and preventing both overpayment and underselling, this is crucial for a fair and efficient real estate market \cite{sayce2009real}. Furthermore, the significance of accurately assessing property values extends deeply into the business arena. Real estate agencies and civil construction enterprises hinge their operations on precise property predictors, pivotal in making well-informed decisions \cite{yang2022graph}.

Traditionally rooted in statistical methodologies, property assessment via automated valuation models (AVMs) has witnessed a transformative shift due to the rapid advancement of deep learning (DL) \cite{tekouabou2024ai}. As such, researchers have redirected their attention toward exploring the application of these cutting-edge models, surpassing the limitations of traditional statistical approaches. In this context, the main virtue of DL models is their capacity to learn spatial pattern effectively using, e.g., recurrent neural networks (RNNs) such as Long Short-Term Memory (LSTM), convolutional neural networks (CNNs), or attention mechanisms \cite{tekouabou2024ai,yang2022graph}.   

This research contributes to the field of house price prediction by showing the efficacy of incorporating geospatial patterns into the modeling process via graph neural networks (GNNs) with transformer convolutional message passing layers. GNNs can capture intricate relationships and dependencies between entities in a graph, making them well-suited for this task. GNN architectures are highly flexible and adaptable to various types of graphs and tasks. GNNs excel at aggregating both local and global information from neighboring nodes in the graph. By iteratively passing messages between nodes, they can capture hierarchical patterns and dependencies, allowing them to make informed predictions \cite{waikhom2023survey}. The main novelty lies in adapting the $k$-nearest similar house sampling (KNHS) approach \cite{bin2019investigations,bin2019peer}, originally designed for LSTM, to the GNN model. This adaptation yields a scalable framework for real estate appraisal, addressing the limitations of the existing GNN approach for automated valuation.

Our main hypothesis asserts that it is possible to improve the current state-of-the-art AVMs by designing novel strategies based on GNNs. Consequently, this hypothesis suggests that exploring innovative deep-learning strategies tailored to the specific challenges of house price prediction can yield superior results compared to traditional approaches. The findings on a propietary dataset of approximately 200,000 houses in the Santiago housing market highlight the potential for improved accuracy and reliability in predicting house prices, providing valuable insights for real estate academics and practitioners, urban planners, and decision-makers in the housing market. Alongside location data, our dataset encompasses a diverse set of features that characterize the properties. Unlike other benchmark datasets reliant on asking prices, i.e., the amount of money the seller desires a buyer to pay \cite{rico2021machine}, ours considers the final selling prices, thereby mitigating potential biases.

The remainder of this paper is structured as follows: prior studies on DL for property assessment are discussed in Section \ref{sec:lit}, formalizing relevant aspects of the paper. Next, the proposed GNN models for automated property valuation are presented in Section \ref{sec:Prop}. Experimental results are reported in Section \ref{sec:Exp}. Finally, Section \ref{sec:Conc} provides the main conclusions of this study, discussing limitations and future extensions.

\section{Prior work on automated valuation models}\label{sec:lit}

There is a vast literature on property assessment models, ranging from naive techniques to state-of-the-art DL methods. The main approaches are discussed next.

\subsection{Comparable sales method}

The comparable sales method is the simplest approach to estimate property prices. This method produces an estimate of the market value for a given property by comparing the transaction prices of similar properties \cite{reEvalKLush}. The market value of these similar properties is used as a point of reference, and the estimated price of a given property is derived based on these references and utilizing some form of correction to account for differences that the property may have with its references. In practice, these corrections are typically made through intuition and domain knowledge rather than a quantitative approach.

\subsection{Hedonic Regression}

Hedonic Regression (HR) models are one of the most widely applied models to estimate the price of a property \cite{HedonicPrices}. This method utilizes a linear regression to measure the contribution of the different property attributes. The basic function of the HR is as follows:
\begin{equation}
V = \alpha_0 + \sum_{i=1}^{n} \alpha_i x_i + \epsilon,
\end{equation}
where $V$ is the estimated market price of a given property, $x_i$ are the property attributes, $\alpha_i$ are the weights assigned to each property attribute, and $\epsilon$ is the error term, for $i=1,...,n$. 

Although the hedonic regression is a more robust approach than the comparable sales method, as it incorporates a more quantitative method and considers more influential factors, the HR model assumes that the relationship between house attributes follows a linear correlation. This assumption is its pitfall, as the literature shows that the market value of a property also depends on nonlinear relationships \cite{alkan2023using,bilgiliouglu2023comparison}.

\subsection{Traditional machine learning models}

The market value of a property depends not only on linear correlations between its features but also on nonlinear relationships. Machine learning models are adept at understanding these nonlinear relationships \cite{alkan2023using,bilgiliouglu2023comparison}.

Regression trees are a popular machine learning technique used for solving regression problems. These models are a type of supervised learning algorithm that partitions the feature space into a hierarchical structure to make predictions of a target variable. The regression tree comprises several nodes and branches. In the context of property valuation, the root node represents the entire dataset of properties. The model then identifies the best property feature and corresponding threshold and divides the data into two subsets. This process repeats for every subset until reaching the allocated depth of the tree. Once the tree construction is complete, each leaf node represents a final prediction value. For property prediction, the prediction at a leaf node is usually the mean value of a target variable within a subset.

The literature shows that regression trees are capable of capturing the nonlinear relationship in property valuation and can achieve superior results to Hedonic Regression. \cite{Kok2017} state that the simplicity and interpretability of these models also make them more suitable than the hedonic model for property assessment. In a study by \cite{dimopoulos2018accuracy}, a random forest outperformed other models, yielding the best results. In contrast, \cite{alkan2023using} reported best results with Support Vector Machines in the context of real estate value prediction in tourism centers. In \cite{park2015using}, best results are achieved with the Ripper Algorithm, a rule-based classification algorithm.

\subsection{Local spatial models}

Although traditional machine learning methods are capable of capturing the nonlinear relationships in property assessment, they often struggle to fully comprehend the spatial relationships that exist between properties. \cite{Kok2017} highlights the influence of various geographical factors, such as distance to points of interest, on property prices. Additionally, \cite{yang2022graph} shows that peer dependence also plays a pivotal role in property pricing.

To incorporate the geospatial impact into property assessment, models need to define the latent influence that exists locally between a given house and its neighboring properties. One widely used model for this purpose is Locally Weighted Regression (LWR). LWR is a sophisticated regression model that aims to estimate property prices by leveraging the spatial dependencies and relationships that may exist within its neighboring properties. LWR assumes that the price of a given property is not only dependent on its characteristics but also on the characteristics of its surrounding properties \cite{huang2010geographically}.

\subsection{Deep Learning approaches for real estate appraisal}

Most of the research conducted in the field of AVMs today focuses on exploring the application of DL techniques to enhance the predictive capabilities of AVM. Models such as RNN \cite{bin2017regression} CNN \cite{poursaeed2018vision}, LSTM \cite{yu2018prediction}, and GNN \cite{yang2022graph} have shown promising results that surpass traditional machine learning models \cite{Abidoye2017ArtificialNN,tekouabou2024ai}.

The Peer Dependence Valuation Model (PDVM), proposed by \cite{bin2019investigations}, aims to utilize the geospatial relationships among properties using deep learning techniques. The PDVM architecture consists of three main components:
\begin{enumerate}
  \item $k$-nearest similar house sampling: This is the initial step of the PDVM architecture. It is employed to aggregate sequences of properties and their neighboring houses.
  \item Bidirectional long short-term memory (B-LSTM) layer: The B-LSTM layer is utilized to process the sequences generated by the KNHS algorithm. It extracts the representation of the to-be-valued house based on the mutual impact of the houses in the sequence.
  \item Fully-connected layer (FC): Finally, an FC serves as a regressor for estimating the house price based on the representations obtained from the previous B-LSTM layers.
\end{enumerate}

Transformers and attention mechanisms have revolutionized natural language processing (NLP) and other sequential data tasks due to their ability to capture long-range dependencies and contextual information more effectively than previous architectures like RNNs and CNNs \cite{vaswani2017attention}.

At the core of transformers is the self-attention mechanism, which enables the model to weigh the importance of different words (or tokens) in a sequence when processing each word. Unlike traditional sequence-to-sequence models like RNNs, transformers can process the entire input sequence in parallel, making them more efficient for long sequences \cite{vaswani2017attention}.

Attention-based DL models have been successfully applied in real estate valuation. For instance, \cite{kucklick2023tackling} utilized a convolutional block attention mechanism for satellite image-based real estate appraisal via CNN. Alternatively, \cite{bin2019attention} developed an attention-based multimodal fusion model for automated valuation.

Finally, another recent trend in DL is GNN, which incorporates graph theory in the modeling process \cite{liu2024heterogeneous}. A graph is a data structure consisting of vertices or nodes ($V$) and edges ($E$). Its structure represents the pairwise relationships between objects. A graph can be defined as follows: $G=(V,E)$. A graph is represented by a matrix, called an adjacency matrix. Given a graph with $N$ nodes and $M$ edges, its corresponding adjacency matrix $A$ will be of dimension $NxN$, and the value at $A[i,j]$ will describe the pairwise relationship that exists between a $V_i$ and $V_j$ \cite{liu2024heterogeneous}.

Graph convolutional networks (GCNs) are well-known DL architectures that perform convolutional operations on graph-structured data \cite{phan2023aspect}. There are two main approaches to convolution: spectral and spatial GCNs, which have distinct advantages and limitations depending on the specific task and characteristics of the graph data. Spectral GCNs operate in the spectral domain, where graph convolution is performed by applying convolutions to the eigenvectors of the graph Laplacian matrix. In contrast, Spatial GCNs operate directly on the spatial domain of the graph, where convolutional operations are performed directly on the nodes and their neighbors in a similar fashion to traditional CNNs \cite{phan2023aspect}.

The spatial regression GCN with external attention (A-SRGCNN) is a model proposed in \cite{yang2022graph}, which utilizes spectral GNNs to predict the prices of properties. It employs a spatial graph structure to map the interactions that exist in the housing market and to mimic a locally weighted regression.

After the output of the graph convolutions, it employs external attention to assign weights to the features of the properties. The resulting embedding is then passed to a linear layer to retrieve the final estimated price of each property.

The addition of a second set of locally learnable weights in the graph convolution allows for defining the importance a certain house has on the predicted house, while the graph structure provides a suitable framework to model global interactions. This, combined with the use of a spectral graph structure, further enhances the spatial interactions by utilizing the spectral properties of the graph to understand global interactions.

However, this architecture has some drawbacks in terms of scalability and predictive performance. The graph structure considers an undirected fully-connected graph, meaning that all the nodes are connected to each other. The resulting graph has a size of $N^2$, where $N$ is the number of nodes in the graph. Implementing the graph convolutions in the spectral domain becomes intractable in terms of running times for large datasets. Notice that spectral GCNs often require computationally intensive operations such as eigen-decomposition of the graph Laplacian matrix, which can be computationally expensive for large graphs \cite{phan2023aspect}.

Additionally, the fully-connected graph structure of A-SRGCNN is counterproductive when modeling the housing market in large areas. Intuitively, the dependency between two houses that are very far away should be very low. Incorporating these patterns can introduce noise to the model, leading to overfitting.


Based on these limitations, this study proposes two novel GNN-based AVM models inspired by the message-passing scheme introduced in \cite{shi2021masked}, which incorporates self-attention in the message-passing layer to understand the spatial interactions in the housing market. The models are implemented using graph convolutions in the spatial domain to address the memory issue associated with the spectral graph structure of A-SRGCNN. In A-SRGCNN, the primary reason for using external attention is to handle the size of the graph, which is too large to utilize self-attention to model pairwise interactions. Since our approaches are based on Spatial GNNs, we can employ self-attention without incurring computationally intensive operations.

\section{The proposed GNN models for automated property valuation}\label{sec:Prop}

In this section, we propose two GNN methods for real estate appraisal. The main idea is to present the housing market data to a deep neural network by structuring it in a way that represents the spatial structure of the dataset, extending the reasoning behind PDVM \cite{bin2019investigations} to spatial GCNs. The main difference between the two proposed GCNs is that the first one utilizes standard graph convolution layers, while the second one employs transformer graph convolutions. Moreover, the latter approach takes into account edge weights, resulting in a more accurate representation of the neighborhood by incorporating distances between properties.

This section is structured as follows: First, the spatial GCNs are formalized in Section \ref{SpatGCN}. Next, the KNHS algorithm developed in \cite{bin2019investigations} is detailed, including the adaptations proposed to improve this approach in the context of GNNs. Finally, the two proposed GCN models for automated property valuation are presented in Section \ref{PD-GCN} and Section \ref{PD-TGCN}. We refer to these approaches as peer dependence - graph convolutional network (PD-GCN) and peer dependence - transformer graph convolutional network (PD-TGCN), respectively.

\subsection{Spatial GCNs}\label{SpatGCN}

The spatial GCN is a type of message passing scheme that operates in the spatial domain, as opposed to spectral GCN. Message passing is a concept used to describe various algorithms on graphs, including those employed in GNNs. It involves exchanging information (messages) between nodes in a graph, typically using a set of update rules \cite{phan2023aspect}.

For each node, information is propagated from its neighboring nodes to create an aggregated representation of each node with its local neighborhood, using convolutional operations.Let $H = [h_1, \ldots, h_n]$ denote the matrix of node features serving as input to a convolutional layer. A graph convolution is defined by combining two operations. First, for each node $v_i$, the information (feature vectors) of its neighbors $N_i = {j:a_{ij} = 1}$ is aggregated \cite{danel2020spatial}:
\begin{equation}
\hat{h}_i = \sum_{j \in N_i} u_{ij} h_j.
\end{equation}
This equation can also be expressed as $\hat{H} = UH^{T}$, where the vector $U$ defines the weights assigned to each connection in the graph. These weights can be either predetermined or trainable. The weights can be predefined as binary values, indicating whether a node is a neighbor of a given node or not, but they can also be defined as scalar values that indicate the interaction between each node.

Finally, a standard MLP is applied to transform the intermediate representation $\hat{H}$ into the final output of a given layer:
\begin{equation}
\text{MLP}(\hat{H};W) = ReLU(W^{T} \hat{H} + b),
\end{equation}
where $W$ is a trainable weight matrix and $b$ is a trainable bias vector \cite{danel2020spatial}.

\subsection{$k$-nearest similar house sampling revisited}\label{KNHS}

Assessing the price of a property involves considering various geospatial characteristics, such as its location and the impact of neighboring properties. To model this phenomenon, the KNHS algorithm creates subgroups of houses that most resemble one another, both geographically and in terms of their features \cite{bin2019investigations}.

The KNHS algorithm operates in two stages. The first stage identifies the geographical neighbors of a given house feature vector $h_i$. This is achieved by computing the orthodromic distance between a property's XY coordinates, typically denoting its main entrance, and those of all other properties within the dataset. It constructs a pairwise distance matrix $A$ and selects the properties that fall within a certain distance threshold $t$, designating them as neighbors. Formally, this can be described as:
\begin{equation}
A = \left[\begin{matrix}
     0 & \sigma_{12} & \sigma_{13} & \ldots & \sigma_{1n}\\
     \sigma_{21} & 0 & \sigma_{23} & \ldots & \sigma_{2n}\\
     \sigma_{31} & \sigma_{32} & 0 & \ldots & \sigma_{3n}\\
     \vdots & \vdots & \vdots & \ddots\\
     \sigma_{n1} & \sigma_{n2} & \sigma_{32} & \ldots & 0
\end{matrix}\right],
\end{equation}
where
\begin{equation} \label{eq:sigma}
\sigma_{ij} = 2\arcsin{\sqrt{\sin^2({\frac{\Delta\phi}{2}}) + \cos({\phi_i}) \cos({\phi_j}) \sin^2({\frac{\Delta\lambda}{2}})} }.
\end{equation}
In Eq. \eqref{eq:sigma}, $\phi_i$ and $\lambda_i$ correspond to the latitude and longitude of $h_i$, respectively. The output of the first stage of the KNHS algorithm will be a matrix $A_{1}^{n\times m}$ containing a sequence for every house $S_i$ containing the feature vectors of all $h_j$ where $\sigma_{ij} < t ;\forall j \in n$.

In the second stage of the KNHS algorithm, the comparison of attribute similarities between a given property and its neighbors takes place. First, the Euclidean distance $d_{ij}$ is used between a given $h_i$ and all $h_j$ in $\hat{S_i}$ to calculate the similarities between a house and all its geospatial neighbors.
\begin{equation} \label{eq:distance}
d_{ij} = \sqrt{\sum_{f=1}^{l}{\left(h_{if} - h_{jf} \right)^2}},
\end{equation}
where $l$ is the number of features. The output sequence of the second stage for a given house $i$ will be a sequence $S_i$ of the $K$ $h_j$ with the smallest $d_{ij}$ from $\hat{S_i}$ including $h_i$. Subsequently, the sequences $S_i = [h_0,\ldots, h_{i-1}, h_i, h_{j+1}, \ldots, h_k]$ are ordered so that $h_i$ is in the middle of the sequence and $h_{i-1}$ and $h_{i+1}$ are the houses most similar to it with the least $d_{ij}$, and the feature vectors $h_0$ and $h_k$ are the most dissimilar houses with the greatest $d_{ij}$.

The original KNHS proposed in \cite{bin2019investigations} mainly emphasizes the significance of positioning $h_i$ in the center of the sequence. It suggests that placing the to-be-valued house in this central position allows the model to capitalize on the advantages of both the forward and backward passes related to the LSTM model. However, it does not consider the order of the remaining elements. In the proposed PD-TGCN method, the use of a graph structure and, in particular, edge attributes and weights, has the advantage of utilizing the order of the neighbors and their distance to $h_i$, incorporating additional information in the modeling process.  

In this revised version of the KNHS algorithm, we consider a weighted variant of the distance function that defines the neighborhood (Eq. \eqref{eq:distance}). The goal is to upweight variables that are more important than others, following the reasoning behind other studies with feature-weighted distance metrics for the computation of $k$-nearest neighbors \cite{maldonado2022fw}. The redefined distance measure follows:

\begin{equation} \label{eq:weighteddistance}
d_{ij} = \sqrt{\sum_{f=1}^{l}{w_f\left(h_{if} - h_{jf} \right)^2}}.
\end{equation}

\subsection{The proposed PD-GCN model}\label{PD-GCN}

The PD-GCN method consists of two stacked graph convolutions using a mean aggregation function. Each graph convolution layer returns a representation of each node based on its local neighborhood information. These representations use trainable weights at each layer of the graph convolution, allowing it to learn the spatial dependencies that exist between each node and its neighboring nodes. Furthermore, by stacking graph convolutions together, the model is able to learn not only the dependencies from its one-hop neighborhood but also from its $n$-hop neighborhood after $n$ stacked graph convolutions, as each layer utilizes the previous representation when aggregating the neighboring nodes' information \cite{morris2021weisfeiler}.

The Graph Convolution block used in the proposed PD-GCN model can be represented through the following expressions. Essentially, each graph convolutional layer updates the feature representation of nodes based on their local neighborhood information, allowing the model to capture and propagate information throughout the graph structure. Let the feature vector $h_i$ of house $i$ before the first convolutional layer be denoted as $h_{i}^{(0)}$. The feature vector of house $i$ computed at the first convolutional layer is as follows:
\begin{equation} \label{eq:hi1}
h_{i}^{(1)} = W^{(0)} h_{i}^{(0)} + W^{(1)} \hat{h}_{i}^{(1)},
\end{equation}
where
\begin{equation} \label{eq:hathi1}
\hat{h}_{i}^{(1)} = ReLU(\frac{\sum_{j \in \mathcal{N}_i} h_{j}^{(0)}}{N}), 
\end{equation}
with $W^{(1)}$, $W^{(0)}$ being the weight matrices, $\mathcal{N}_i$ representing the set of neighbors of house $i$, $N$ the number of neighbors of house $i$, and $h_{j}^{(0)}$ the feature vector of house $j$. The second convolutional layer has the following form:
\begin{equation} \label{eq:hi2}
h_{i}^{(2)} = W^{(1)} h_{i}^{(1)} + W^{(2)} \hat{h}_{i}^{(2)},
\end{equation}
where
\begin{equation} \label{eq:hathi2}
\hat{h}_{i}^{(2)} = ReLU(\frac{\sum_{j \in \mathcal{N}_i} h_{j}^{(1)}}{N}).
\end{equation}

The proposed PD-GCN architecture is depicted in Figure \ref{GNN-GCN-diagram}. First, the input feature vectors are split into two parts: one containing the continuous features and the other containing the encoded categorical features. The categorical features are then processed through an embedding layer. Subsequently, the categorical embeddings are concatenated back with the continuous features and passed to the GCN block. 

The GCN block is the central component of the model and comprises two stacked GCN layers with a mean aggregation function. Each GCN layer extracts the representation of a given house along with its local neighborhood by aggregating the neighborhood information of each house using learnable weights. By stacking two GCN layers, the model not only learns the dependencies within its immediate one-hop neighborhood but also extends its learning to the neighborhoods of its neighbors (two-hop neighborhood). This capability allows the model to capture the complex interactions within the large cities.

The resulting node embeddings from the second convolutional layer are then fed through a dense layer with a ReLU activation function and finally passed through a dense linear layer to obtain the predicted prices of the houses.

\subsection{Peer Dependence Transformer Graph Convolution Network (PD-TGCN)}\label{PD-TGCN}

The proposed PD-TGCN model is an adaptation for real estate appraisal of the TGCN model introduced in \cite{shi2021masked}. The main concept behind the TGCN is to represent interactions between nodes using attention mechanisms, instead of relying on traditional convolutional operations. The self-attention mechanism utilized by the TGCN comprises three components: query, key, and value. The feature vector of each node undergoes three linear transformations to produce query, key, and value vectors. Subsequently, attention scores between nodes are computed by taking the dot product between the query and key vectors, which are then normalized using a softmax operation. Utilizing these attention scores, the TGCN calculates weighted sums of the value vectors to derive the node representations. Through the stacking of multiple TGCN layers, the model is capable of capturing intricate and high-order interactions within the graph structure.

In the proposed PD-TGCN model, the two transformer graph convolutions can be elucidated through the following steps:

\noindent \textbf{First Transformer Convolution Layer:}
    \begin{itemize}
        \item \textbf{Input:}
            The feature vectors of the houses, denoted as 
            $H^{(0)} = [h_1^{(0)}, h_2^{(0)}, ..., h_N^{(0)}]$, and the attribute vectors of the edges, denoted as $E = [e_{12}, e_{14}, ... e_{nm}]$
        \item \textbf{Query, Key, and Value}:   
            The feature vectors are transformed into query $(Q^{(0)})$, key $(K^(0))$, and value $(V^{(0)})$ vectors. 
            The $Q^{(0)}$, $K^{(0)}$, and $V^{(0)}$ vectors are obtained by applying linear transformations to the input feature vectors. A fourth vector $U^{(0)}$ is obtained by an additional linear transformation applied to the edge attribute vector. Formally, for a house $i$, the $Q^{(0)}$, $K^{(0)}$, and $V^{(0)}$ vectors are: 
 \begin{equation} \label{eq:QKV}
            Q_i^{(0)} = \text{linear}_{\text{query}}(h_i^{(0)}), \ K_i^{(0)} = \text{linear}_{\text{key}}(h_i^{(0)}), \ V_i^{(0)} = \text{linear}_{\text{value}}(h_i^{(0)}).
\end{equation}
            Similarly for an edge between house $i$ and its neighbor $j$, the $U^{(0)}$ vector is:
 \begin{equation} \label{eq:edge}
U_{ij}^{(0)} = \text{linear}_{\text{edge}}(e_{ij}).
\end{equation}
        \item \textbf{Self-Attention}:
            The convolutional layer then calculates the self-attention scores using the $Q^{(0)}$, $K^{(0)}$ and $U^{(0)}$ vectors. For each house $i$, the attention score between house $i$ and its neighbor $j$ is computed as the dot product of their $Q^{(0)}$ and $(K^{(0)}+U^{(0)})$ vectors:
 \begin{equation} \label{eq:score}
            \text{AttScore}(i, j)^{(0)} =\frac{{Q_i^{(0)} \cdot (K_j^{(0)}+U_{ij}^{(0)})}}{\sqrt{d_k^{(0)}}},
\end{equation}
where $d_k^{(0)}$ represents the dimension of $K_j^{(0)}$ and is used to offset the increasing magnitude of the dot products as the dimension grows. The attention weight is then calculated by using a softmax function over all the attention scores of the neighbors of node $i$, transforming the attention scores into a probability distribution where the sum of the attention weights of the neighbors is equal to 1. This scaling is done to determine how much information should be propagated from each neighboring house $j$ to house $i$.
 \begin{equation} \label{eq:weights}
             \text{AttWeight}(i, j)^{(0)} = \frac{\text{exp}(\text{AttScore}^{(0)}(i, j))}{\sum_{k \in \mathcal{N}_i} \text{exp}(\text{AttScore}(i, k)^{(0)})}.
\end{equation}   
        \item \textbf{Weighted sum (aggregation):}  
            The feature vectors of the neighboring houses and the edge attributes between the house $i$ and its neighbors are then aggregated using the attention scores. For each house $i$, the aggregated feature vector is obtained by taking the weighted sum of the $V$ vectors of the neighboring houses combined with the edge attributes $U$, where the attention scores act as the weights: 
  \begin{equation} \label{eq:weightsum}
            \hat{h}_i^{(0)} = \sum_{j \in \mathcal{N}_i} \text{AttWeight}(i, j)^{(0)} \cdot (V_j^{(0)} + U_{ij}^{(0)}).
\end{equation}             
        \item \textbf{Linear transformation:}  
            Finally, the aggregated feature vector containing the information of house $i$ local neighborhood is combined with the original input feature vector of house $i$ $h^(0)$ using a linear transformation: 
  \begin{equation} \label{eq:lineartrans}
{h}_i^{(1)} = \text{linear}_{\text{aggr}}([h_i^{(0)}, \hat{h}_i^{(0)}])
\end{equation}     
           where $\text{linear}_{\text{aggr}}$ is a linear transformation applied to the concatenated input and aggregated feature vectors. The edge attributes are not altered in a transformer graph convolution, in the purpose of maintaining the structural information of the graph.
           
            The output of the first Transformer Convolution Layer is a feature vector of all the house aggregated with its one-hop neighbor features denoted as $H^{(1)} = [h_1^{(1)}, h_2^{(1)}, ..., h_N^{(1)}]$.
    \end{itemize}

\noindent \textbf{Second Transformer Convolution Layer:}
    \begin{itemize}
        \item Input:
            The output feature vectors from the first layer, containing all nodes aggregated with their one-hop neighbors $H^{(1)}$ and the original edge attributes of the graph.
        \item Query, Key, and Value:  
            Similar to the first layer, the output feature vectors ${H}^{(1)}$ and the edge attribute vectors are transformed into $Q$, $K$, $V$ and $U$ vectors. 
        \item Self-Attention:
            The self-attention scores and weights are calculated the same way as the  steps are performed similarly to the first layer, using the $Q$, $K$, $V$ and $U$ vectors.
        \item Weighted Sum (Aggregation):
            The feature vectors of the neighboring nodes are then aggregated using the attention scores. For each node $i$, the aggregated feature vector is obtained by taking the weighted sum of the $V$ vectors of the neighboring nodes, where the attention scores act as the weights:
  \begin{equation} \label{eq:weightsum2}
            \hat{h}_i^{(1)} = \sum_{j \in \mathcal{N}_i} \text{Attention Weight}(i, j)^{(1)} \cdot (V_j^{(1)} + U_{ij}^{(1)})
\end{equation}  
        \item Linear Transformation:
           Finally, the aggregated feature vector containing the information of house $i$ two-hop neighborhood is combined with the representation of house $i$ one-hop neighborhood feature vector using a linear transformation.
  \begin{equation} \label{eq:lineartrans2}
{h}_i^{(2)} = \text{linear}_{\text{aggr}}([h_i^{(1)}, \hat{h}_i^{(1)}]).
\end{equation}  
           
            The output feature of the second transformer convolution layer is a feature vector of all the house aggregated with its two-hop neighbor features denoted as 
            $H^{(2)} = [h_1^{(2)}, h_2^{(2)}, ..., h_N^{(2)}]$.
    \end{itemize}

The proposed PD-TGCN architecture is depicted in Figure \ref{GNN-TGCN-diagram}. The model utilizes an embedding layer to train the representations of the categorical features. The output of the embedding layer is then concatenated with continuous features. These concatenated feature vectors are then separately passed with the edge attributes to a Transformer Convolution (TC) block. The TC block contains two stacked TC layers. The first convolution creates a representation of all houses within their one-hop neighborhoods, and by applying a second convolutional layer, the model creates a representation of the two-hop neighborhood as well. Finally, the result from the second TC layer is passed through a linear layer to extract the house prices.

\begin{landscape}
\begin{figure}[ht!]
\caption{PD-GCN diagram}
\centering
\label{GNN-GCN-diagram}
\includegraphics[width=.99\linewidth]{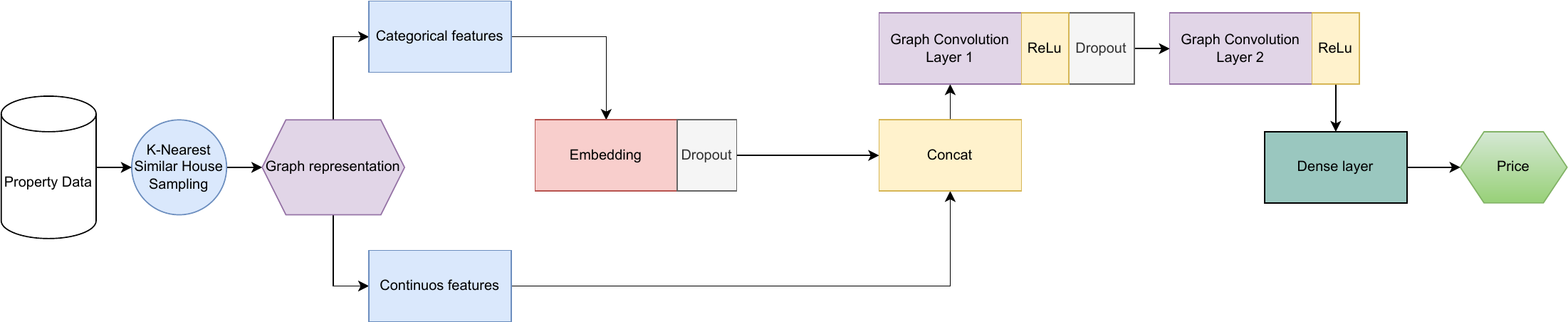}
\end{figure}

\begin{figure}[ht!]
\caption{PD-TGCN diagram}
\centering
\label{GNN-TGCN-diagram}
\includegraphics[width=.99\linewidth]{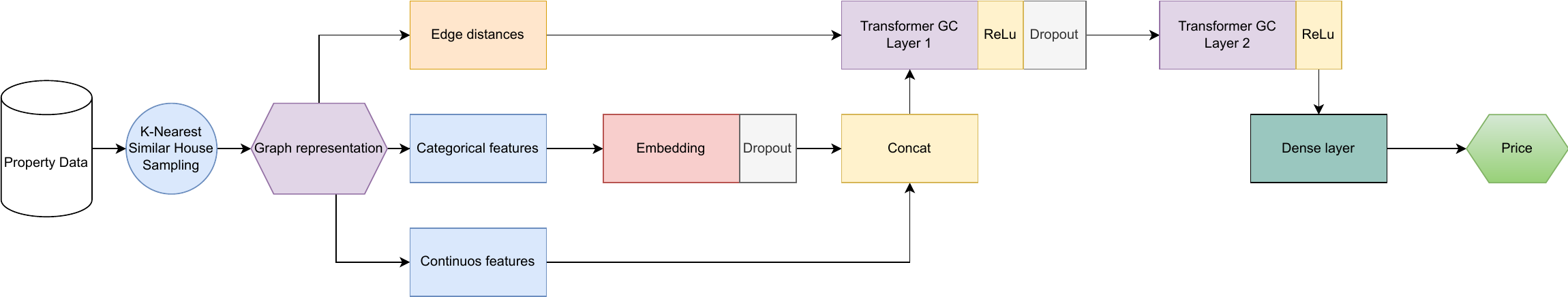}
\end{figure}
\end{landscape}

Each transformer graph convolution layer also employs a multi-head attention mechanism. Each head in the transformer model has its own set of query ($Q$), key ($K$), and value ($V$) transformations. This allows each head to focus on different patterns and dependencies within the data, capturing diverse and complementary information.

For both models, two graph convolution layers were used because of how GNNs work: in each message passing layer, information from neighboring nodes is added to each node, so after one layer, each node will result in the aggregation of its features with those of its first-order neighbors. In the second layer, each node now not only represents its own information but also that of its neighborhood, which then provides a more complete picture of the spatial interactions that exist in the graph. However, including more than two layers leads to the over-smoothing problem. When this phenomenon occurs, information propagated across multiple layers becomes excessively smoothed or averaged out, leading to the loss of important details and nuances in the data \cite{chen2023agnn}. Empirically, we observed that adding more than two layers resulted in worse performance for the model, despite the use of regularization to try to mitigate the effect of over-smoothing.

\section{Experimental Results}\label{sec:Exp}

We applied the proposed GNN models to a proprietary dataset comprising transaction prices of houses in Santiago, Chile. This section is divided as follows: Section \ref{sec:Sett} describes the experimental setting, while a summary of the results is presented in Section \ref{sec:Summ}. Finally, results of a sensitivity analysis for the main parameters are reported in Section \ref{sec:Sens}.

\subsection{Experimental setting and dataset}\label{sec:Sett}

The final dataset consists of 225,000 houses described by 50 explanatory variables. The data was collected in the 2009-2019 period. All prices are given in UF (\emph{unidad de fomento}), which serves as the unit of account. Its exchange rate with the Chilean peso is regularly recalibrated to counter inflation, thereby maintaining the purchasing power of the UF nearly steady on a day-to-day basis. This results in a more reliable estimation of a property values in different time periods. The UF value was adjusted semiannually. As a reference, 1 UF = 38.07 USD at 04/17/2024.

The following exploratory variables, we considered for modelling purposes:
\begin{itemize}
\item Seven numerical variables describing the property, including its surface area and the appraisal value for taxation purposes.
\item Seventeen numerical variables describing the district in which the property is located, such as the number of inhabitants, the number of houses in the district or the total area of the aggregated houses' lot area in that district.
\item Three categorical variables describing both the property and the district, such as the main building material used in construction.
\item Twenty-three numerical variables indicating the distance from the property to points of interest, such as subway stations, high schools, shopping centers, bus stops, or hospitals. These variables have shown to be useful in previous studies \cite{DACCI201971}.
\end{itemize}

The following exclusion criteria were utilized in order to filter out noisy samples:
\begin{itemize}
\item The appraisal for taxation purposes tends to be around 30\% of the market value. Following an investigation from the Chilean central bank designed to identify inconsistencies \cite{Flores_2018}, only properties where the ratio between the appraisal for taxation purposes and the transaction price is less than 10 and greater than 1 are kept. 3.4\% of the samples were deleted in this step.
\item Transaction prices outside of the expected range were deleted based on domain knowledge of the Santiago housing market (0.03\% of the samples). All values greater than 50,000 UF and less than 400 UF were discarded.
\item Only prices per m$^2$ less than 6000 UF and greater than 30 are kept. This range was created based on expert knowledge in the field, discarding 0.03\% of the samples.
\item Properties sold multiple times in one year were considered market distortions and were therefore discarded (2.3\% of the samples).
\item All houses containing coordinates that are inconsistent with the commune were also deleted (0.002\% of the samples). A commune is the smallest administrative subdivision in Chile, and the province of Santiago encompasses 32 communes.
\end{itemize}

In order to perform an in-depth analysis of the housing market, eight additional datasets were created by defining groups of adjacent communes. The sample size of each group is reported in Table \ref{tab:comunes}.

\begin{table}[ht]
\caption{Groups of adjacent communes of Santiago, Chile.}
    \label{tab:comunes}
\centering
\resizebox{\textwidth}{!}{
\begin{tabular}{ccc}
\toprule
\textbf{Group} & \textbf{Comunes} & \textbf{Rows} \\
\midrule
Group 1 & LO BARNECHEA, LAS CONDES, VITACURA, PROVIDENCIA & 14,799 \\
Group 2 & NUNOA, LA REINA, MACUL, PENALOLEN & 15,489 \\
Group 3 & LA FLORIDA, PUENTE ALTO, LA PINTANA, LA GRANJA & 51,453 \\
Group 4 & SAN RAMON, LA CISTERNA, SAN MIGUEL, SAN JOAQUIN, PEDRO AGUIRRE CERDA & 8268\\
Group 5 & SAN BERNARDO, EL BOSQUE, LO ESPEJO, CERRILLOS & 19,904 \\
Group 6 & ESTACION CENTRAL, MAIPU, QUINTA NORMAL, LO PRADO, SANTIAGO & 36,956\\
Group 7 & PUDAHUEL, CERRO NAVIA, RENCA, QUILICURA & 25,661 \\
Group 8 & HUECHURABA, INDEPENDENCIA, RECOLETA, CONCHALI & 9056 \\
\bottomrule
\end{tabular}}
\end{table}

The datasets were split into training and testing sets with a ratio of 75:25. The mean squared error (MSE) is used as the loss function for all models. It measures the average squared difference between the predicted and actual values. For model evaluation, we consider the mean absolute percentage error (MAPE), which is a widely used metric to assess the accuracy of predictions in property valuation models. It measures the percentage difference between the actual property values and the corresponding model predictions. This makes it easily interpretable in the context of house prices and quantifies how far off the model predictions are from the actual prices on average, in percentage terms.

Regarding variable transformations, the Box-Cox mapping was considered to correct skewness (logarithmic transformation). The min-max scaling was performed on the continuous features. For nominal variables, we use categorical embeddings, as suggested in \cite{suenaga_2018}. This strategy has shown better results than one-hot encoding in deep learning as it captures relationships between categories, being also more efficient in terms of computational costs \cite{suenaga_2018}. As a rule of thumb, the sizes of these categorical embeddings are given by:

  \begin{equation} \label{eq:embsize}
\text{emb size} = (N_{\text{categories}}, \textbf{min}( 50, ( \frac{N_{\text{categories}} + 1}{2})
\end{equation}

\subsection{Results summary}\label{sec:Summ}

We consider Moran's I \cite{moran1950notes} for a preliminary geo-spatial analysis. This metric quantifies the degree of similarity between neighboring observations in a spatial dataset. A Moran's I value ranges from -1 to +1. A positive value indicates positive spatial auto-correlation (i.e., similar values are clustered together), a negative value indicates negative spatial auto-correlation (i.e., similar values are dispersed), and a value close to zero indicates no spatial auto-correlation \cite{moran1950notes}.

The Moran's I measure for the Santiago housing market is 0.81 with a p-value of 0.001. This indicates a strong positive spatial auto-correlation, implying that properties located near each other tend to have similar prices. The low p-value suggests that the observed Moran's I value is statistically significant. This result is important as it suggests that models that can accurately model spatial patterns should achieve the best predictive results.

Next, the results in terms of MAPE are presented in Table \ref{tab:MAPEallModelsGroupAll}. We compare the proposed PD-GCN and PD-TGCN models with a standard linear regression (LINREG), Random Forest (RF) and Extreme Gradient Boosting (XGBoost), and PDVM as a DL model tailored for real estate appraisal.

\begin{table}[ht]
\caption{MAPE of all models trained in different subsets of adjacent communes. The results are highlighted using color coding: the best predictive performance is emphasized with a dark green color.}
\label{tab:MAPEallModelsGroupAll}
\centering
\resizebox{\textwidth}{!}{
\begin{tabular}{lrrrrrrrrr}
\toprule
Method & All & Group$_1$ & Group$_2$ & Group$_3$ & Group$_4$ & Group$_5$ & Group$_6$ & Group$_7$ & Group$_8$ \\
\midrule
LINREG & {\cellcolor[HTML]{A50026}} \color[HTML]{F1F1F1} 0.246 & {\cellcolor[HTML]{A50026}} \color[HTML]{F1F1F1} 0.213 & {\cellcolor[HTML]{A50026}} \color[HTML]{F1F1F1} 0.283 & {\cellcolor[HTML]{A50026}} \color[HTML]{F1F1F1} 0.233 & {\cellcolor[HTML]{A50026}} \color[HTML]{F1F1F1} 0.312 & {\cellcolor[HTML]{A50026}} \color[HTML]{F1F1F1} 0.210 & {\cellcolor[HTML]{A50026}} \color[HTML]{F1F1F1} 0.228 & {\cellcolor[HTML]{A50026}} \color[HTML]{F1F1F1} 0.204 & {\cellcolor[HTML]{CC2627}} \color[HTML]{F1F1F1} 0.288 \\
RF & {\cellcolor[HTML]{128A49}} \color[HTML]{F1F1F1} 0.207 & \bfseries {\cellcolor[HTML]{006837}} \color[HTML]{F1F1F1} 0.198 & {\cellcolor[HTML]{6BBF64}} \color[HTML]{000000} 0.264 & \bfseries {\cellcolor[HTML]{006837}} \color[HTML]{F1F1F1} 0.196 & {\cellcolor[HTML]{ED5F3C}} \color[HTML]{F1F1F1} 0.309 & \bfseries {\cellcolor[HTML]{006837}} \color[HTML]{F1F1F1} 0.169 & \bfseries {\cellcolor[HTML]{006837}} \color[HTML]{F1F1F1} 0.214 & {\cellcolor[HTML]{097940}} \color[HTML]{F1F1F1} 0.178 & {\cellcolor[HTML]{FDB768}} \color[HTML]{000000} 0.282 \\
XGBoost & {\cellcolor[HTML]{5DB961}} \color[HTML]{F1F1F1} 0.212 & {\cellcolor[HTML]{D9EF8B}} \color[HTML]{000000} 0.204 & {\cellcolor[HTML]{A0D669}} \color[HTML]{000000} 0.266 & {\cellcolor[HTML]{48AE5C}} \color[HTML]{F1F1F1} 0.202 & {\cellcolor[HTML]{C21C27}} \color[HTML]{F1F1F1} 0.311 & {\cellcolor[HTML]{4EB15D}} \color[HTML]{F1F1F1} 0.176 & {\cellcolor[HTML]{6EC064}} \color[HTML]{000000} 0.217 & {\cellcolor[HTML]{5AB760}} \color[HTML]{F1F1F1} 0.182 & {\cellcolor[HTML]{A50026}} \color[HTML]{F1F1F1} 0.290 \\
PDVM & {\cellcolor[HTML]{4BB05C}} \color[HTML]{F1F1F1} 0.211 & {\cellcolor[HTML]{FDBF6F}} \color[HTML]{000000} 0.208 & {\cellcolor[HTML]{FED27F}} \color[HTML]{000000} 0.274 & {\cellcolor[HTML]{93D168}} \color[HTML]{000000} 0.206 & {\cellcolor[HTML]{54B45F}} \color[HTML]{F1F1F1} 0.298 & {\cellcolor[HTML]{FEE593}} \color[HTML]{000000} 0.193 & {\cellcolor[HTML]{FFFEBE}} \color[HTML]{000000} 0.221 & {\cellcolor[HTML]{E9F6A1}} \color[HTML]{000000} 0.189 & {\cellcolor[HTML]{FEEC9F}} \color[HTML]{000000} 0.279 \\
PD-GCN & {\cellcolor[HTML]{0C7F43}} \color[HTML]{F1F1F1} 0.206 & {\cellcolor[HTML]{91D068}} \color[HTML]{000000} 0.202 & {\cellcolor[HTML]{4BB05C}} \color[HTML]{F1F1F1} 0.263 & {\cellcolor[HTML]{48AE5C}} \color[HTML]{F1F1F1} 0.202 & \bfseries {\cellcolor[HTML]{006837}} \color[HTML]{F1F1F1} 0.295 & {\cellcolor[HTML]{73C264}} \color[HTML]{000000} 0.178 & {\cellcolor[HTML]{9DD569}} \color[HTML]{000000} 0.218 & {\cellcolor[HTML]{219C52}} \color[HTML]{F1F1F1} 0.180 & \bfseries {\cellcolor[HTML]{006837}} \color[HTML]{F1F1F1} 0.265 \\
PD-TGCN & \bfseries {\cellcolor[HTML]{006837}} \color[HTML]{F1F1F1} 0.204 & \bfseries {\cellcolor[HTML]{006837}} \color[HTML]{F1F1F1} 0.198 & \bfseries {\cellcolor[HTML]{006837}} \color[HTML]{F1F1F1} 0.259 & {\cellcolor[HTML]{06733D}} \color[HTML]{F1F1F1} 0.197 & {\cellcolor[HTML]{DC3B2C}} \color[HTML]{F1F1F1} 0.310 & {\cellcolor[HTML]{18954F}} \color[HTML]{F1F1F1} 0.173 & {\cellcolor[HTML]{128A49}} \color[HTML]{F1F1F1} 0.215 & \bfseries {\cellcolor[HTML]{006837}} \color[HTML]{F1F1F1} 0.177 & {\cellcolor[HTML]{0A7B41}} \color[HTML]{F1F1F1} 0.266 \\
\bottomrule
\end{tabular}}
\end{table}

In Table \ref{tab:MAPEallModelsGroupAll}, we observe generally good results, with the best MAPE around 20\% for most datasets. Among the nine datasets, the proposed models achieve the best results in six, while RF obtained the lowest MAPE in the remaining three datasets. Specifically, PD-GCN and PD-TGCN achieve the best results in two and four datasets, respectively. The GNN models outperform PDVM, which is the main benchmark in this context.

For the main dataset that encompasses all houses (second column in Table \ref{tab:MAPEallModelsGroupAll}), the best methods are the proposed PD-TGCN and PD-GCN. This result emphasizes their ability to deal with heterogeneous prices in different areas of the city by defining suitable neighborhoods.

Traditional machine learning models show good results overall, with RF being the third-best model behind the proposed approaches in terms of average MAPE. This can be due to the inclusion of spatial variables in the feature engineering process, such as distances to points of interest.

\begin{figure}[ht!]
\caption{MAPE by comune for the PD-TGCN method.}
\centering
\label{MAPEbycomunePD-TGCN}
\includegraphics[scale=0.4]{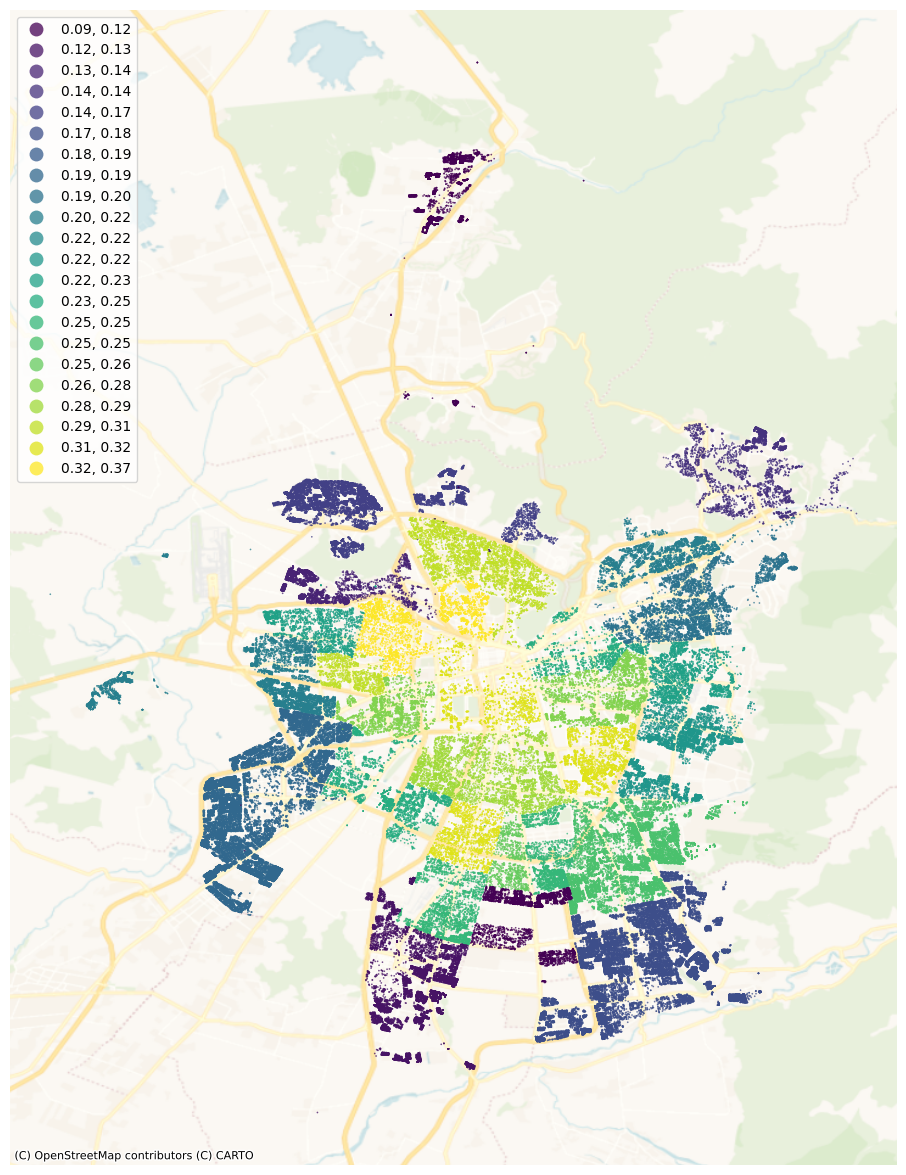}
\end{figure}

Figure \ref{MAPEbycomunePD-TGCN} illustrates the distribution of the MAPE measure at the commune level for the PD-TGCN model, which achieved the lowest error. The colors on the map represent the various MAPE values for each commune.

We observe in Figure \ref{MAPEbycomunePD-TGCN} that the communes located in the city center exhibit the poorest predictive results, whereas the model demonstrates greater efficiency in forecasting property prices for communes situated on the outskirts of Santiago. This trend is consistent across all models. One plausible explanation for this phenomenon is the higher concentration of apartment buildings in the central region of Santiago. Since the data used in this study is limited to houses, the methods cannot fully model the interactions between houses and neighboring apartment buildings. Another source of noise stems from the sale of several downtown houses during this period for the purpose of demolition and subsequent construction of buildings. This activity has contributed to distorting the real estate market in the area.

\subsection{Sensitivity analysis}\label{sec:Sens}

In this section, we discuss the influence of three different variants and parameter values associated with the proposed model. The results are presented in tables \ref{tab:PDGCNExperiments} and \ref{tab:PDTCExperiments} in \ref{sec:sensTables}. In these tables, the metrics $R^2$, Root Mean Squared Error (RMSE), and MAPE are reported in the second, third, and fourth columns, respectively. RMSE is the square root of the MSE measure, while $R^2$ computes the proportion of the variance in the target variable that is predictable from the model.

First, we considered different scaling weights for the appraisal for taxation purposes when calculating the Euclidean distance between two houses in the KNHS algorithm. This was done to investigate how the definition of a more similar house impacts the results of different models. The choice to up-weigh this variable was made based on its demonstrated high correlation with home prices. We explore $w \in {1.2,1.5,1.8,3}$; see the fifth column in tables \ref{tab:PDGCNExperiments} and \ref{tab:PDTCExperiments}.

The results show that the measures are only slightly affected by this scaling factor. The best results are obtained with $w=3$ and $k=16$, showing that up-weighting the appraisal for taxation purposes has a positive effect in terms of predictive performance.

For the second set of experiments, we explore different values for $k$, the neighbors in the KNHS algorithm, to understand how this parameter influences the predictions of the various models. We explore $k \in {6,8,10,12,14,16}$; see the sixth column in tables \ref{tab:PDGCNExperiments} and \ref{tab:PDTCExperiments}.

In these tables, we observe that greater $k$ values improve the results of the KNHS algorithm, achieving the best performance with $k=16$. We conclude that having a large neighborhood has its benefits in terms of performance; however, a negative consequence is a larger computational cost.

In the last experiment, variants of the KNHS algorithm were tested to evaluate different approaches to connecting neighboring houses. In particular, we compare the standard KNHS algorithm (normal) against random ordering and geospatial k-nearest neighbors (geo); see the seventh column in tables \ref{tab:PDGCNExperiments} and \ref{tab:PDTCExperiments}.

Regarding this third experiment, results show that the KNHS algorithm does improve the model results for all three metrics. This indicates that the KNHS algorithm provides important information that the model is able to train upon to understand the Santiago housing market at a deeper level.

The KNHS algorithm was also compared with a geospatial ordering to select the $k$ nearest house. The resulting predictions when utilizing this strategy worsen the predictive power for all three metrics when compared with the original KNHS. From this result, we can conclude that the geospatial interactions do not provide sufficient information for the model to improve its understanding of the housing market, and that the second step of the KNHS algorithm of selecting the houses most correlated with each other plays an important role in the model effectiveness.

\section{Conclusions}\label{sec:Conc}

In this study, two graph-based DL models are proposed to predict property prices by leveraging the geospatial interactions that exist between properties. We adapt the KNHS algorithm to model spatial patterns based on neighboring houses, taking advantage of the structure of the spatial GCN model.

The main difference between the two proposed architectures lies in the approach to perform graph convolutions. The PD-GCN method utilizes two standard graph convolutions using a mean aggregation function. In contrast, the PD-TGCN model considers two stacked transformer convolution layers.

Our experimental results on a proprietary dataset for house price prediction in Santiago, Chile, confirm the virtues of the proposed models. The PD-TGCN achieves the best overall results for the nine different sets of experiments designed based on groups of adjacent communes. This method also performs best when all the houses in the sample were considered. The PDVM model, an LSTM-based approach tailored for real estate appraisal, was outperformed by PD-TGCN.

The improvement of the PD-TGCN model when compared with the PD-GCN model also provides insight into the local relationships that exist within a neighborhood. Its improvement originates from the use of the attention mechanism, showcasing that not all neighboring properties affect a given house equally, and understanding these interactions improves the accuracy of the predicted property's price.

Based on our results, our main hypothesis that it is possible to improve the current state-of-the-art AVMs by designing novel strategies based on GNNs holds. With a 20.4\% MAPE, we can also confirm that it is possible to obtain accurate property value evaluation in the Santiago housing market using state-of-the-art artificial intelligence techniques. As a reference, \cite{bin2019peer} reported a MAPE of 17.5\% for the city of Los Angeles using DL, while the best DL models in \cite{bin2019investigations} achieved 20.12\% and 29.24\% for the cities of Chicago and Detroit, respectively. Our results also confirm the existence of geo-spatial dependencies within the Santiago housing market.

The findings of this study offer promising implications for the broader application of artificial intelligence in the real estate industry and emphasize the need for continual research and development in this area. One limitation of the proposed approach is the use of unimodal models, while the current trend in DL is multimodal learning and information fusion. Multimodal models for real estate appraisal have shown excellent results by, for example, including geographical presentation from street maps in the form of images. Multimodal graph networks are a promising opportunity for future research.

\bibliographystyle{IEEEtran}
\bibliography{biblioAVM}

\newpage

\appendix
\renewcommand\theequation{A.\arabic{equation}}
\setcounter{table}{0} 

\section{Abbreviations}\label{sec:abbreviations}

\noindent  ASRGCNN: Spacial Regression GCN withh external attention \\
AVM: Automated Valuation Models\\
BLSTM: Bi-directional Long Short Term Memory \\
CNNs: Convolutional Neural Networks \\
DL: Deep Learning \\
FC: Fully Connected \\
GCN: Graph Convolution Network \\
GDP: Gross Domestic Product \\
GNN: Graph Neural Network \\
HR: Hedonic Regression \\
KNHS: K-Nearest House Sampling \\
LINREG: Linear Regression \\
LSTM: Long Short-Term Memory \\
LWR: Locally Weighted Regression \\
MAPE: Mean Absolute Percentage Error \\
MLP: Multilayer perceptron \\
MSE: Mean Squared Error \\
NLP: Natural Lenguaje Processing \\
PD-GCN: Peer dependence graph convolutional network \\
PD-TGCN: Peer dependence transformer graph convolutional network \\
PDVM: Peer Dependence Valuation Model \\
RF: Random Forest \\
RMSE: Root Mean Squared Error \\
RNNs: Recurrent Neural Networks \\
TGCN: Transformer Graph Convolutional Network \\
UF:\emph{Unidad de Fomento} (Chilean unit of account) \\
XGBoost: Extreme Gradient Boosting
\newpage

\section{Tables associated to the sensitivity analysis}\label{sec:sensTables}

\begin{table}[ht!]
\centering
\caption{Experiment using PD-GCN}
\label{tab:PDGCNExperiments}
\begin{tabular}{lrrrrrl}
\toprule
 & R2 & RMSE & MAPE & weight & k & knhs \\
\midrule
0 & {\cellcolor[HTML]{A2D76A}} \color[HTML]{000000} 0.923 & {\cellcolor[HTML]{78C565}} \color[HTML]{000000} 788.868 & {\cellcolor[HTML]{D5ED88}} \color[HTML]{000000} 0.210 & {\cellcolor[HTML]{A50026}} \color[HTML]{F1F1F1} 3.000 & \bfseries {\cellcolor[HTML]{006837}} \color[HTML]{F1F1F1} 6 & \bfseries normal \\
5 & {\cellcolor[HTML]{0B7D42}} \color[HTML]{F1F1F1} 0.926 & \bfseries {\cellcolor[HTML]{006837}} \color[HTML]{F1F1F1} 773.212 & {\cellcolor[HTML]{006837}} \color[HTML]{F1F1F1} 0.206 & {\cellcolor[HTML]{A50026}} \color[HTML]{F1F1F1} 3.000 & {\cellcolor[HTML]{A50026}} \color[HTML]{F1F1F1} 16 & \bfseries normal \\
10 & {\cellcolor[HTML]{C9E881}} \color[HTML]{000000} 0.922 & {\cellcolor[HTML]{BDE379}} \color[HTML]{000000} 796.935 & {\cellcolor[HTML]{4BB05C}} \color[HTML]{F1F1F1} 0.208 & {\cellcolor[HTML]{FFFEBE}} \color[HTML]{000000} 1.500 & {\cellcolor[HTML]{66BD63}} \color[HTML]{F1F1F1} 8 & \bfseries normal \\
15 & {\cellcolor[HTML]{D5ED88}} \color[HTML]{000000} 0.921 & {\cellcolor[HTML]{CBE982}} \color[HTML]{000000} 798.698 & {\cellcolor[HTML]{66BD63}} \color[HTML]{F1F1F1} 0.208 & {\cellcolor[HTML]{D9EF8B}} \color[HTML]{000000} 1.200 & {\cellcolor[HTML]{66BD63}} \color[HTML]{F1F1F1} 8 & \bfseries normal \\
20 & {\cellcolor[HTML]{98D368}} \color[HTML]{000000} 0.923 & {\cellcolor[HTML]{6BBF64}} \color[HTML]{000000} 787.499 & {\cellcolor[HTML]{0B7D42}} \color[HTML]{F1F1F1} 0.207 & {\cellcolor[HTML]{A50026}} \color[HTML]{F1F1F1} 3.000 & {\cellcolor[HTML]{FEE08B}} \color[HTML]{000000} 12 & \bfseries normal \\
25 & {\cellcolor[HTML]{E5F49B}} \color[HTML]{000000} 0.921 & {\cellcolor[HTML]{DFF293}} \color[HTML]{000000} 801.659 & {\cellcolor[HTML]{70C164}} \color[HTML]{000000} 0.208 & {\cellcolor[HTML]{A50026}} \color[HTML]{F1F1F1} 3.000 & {\cellcolor[HTML]{66BD63}} \color[HTML]{F1F1F1} 8 & \bfseries normal \\
30 & {\cellcolor[HTML]{2DA155}} \color[HTML]{F1F1F1} 0.925 & {\cellcolor[HTML]{16914D}} \color[HTML]{F1F1F1} 779.178 & {\cellcolor[HTML]{138C4A}} \color[HTML]{F1F1F1} 0.207 & {\cellcolor[HTML]{FEE08B}} \color[HTML]{000000} 1.800 & {\cellcolor[HTML]{A50026}} \color[HTML]{F1F1F1} 16 & \bfseries normal \\
35 & {\cellcolor[HTML]{48AE5C}} \color[HTML]{F1F1F1} 0.925 & {\cellcolor[HTML]{EBF7A3}} \color[HTML]{000000} 803.784 & {\cellcolor[HTML]{6BBF64}} \color[HTML]{000000} 0.208 & {\cellcolor[HTML]{FEE08B}} \color[HTML]{000000} 1.800 & {\cellcolor[HTML]{F46D43}} \color[HTML]{F1F1F1} 14 & \bfseries normal \\
40 & {\cellcolor[HTML]{ADDC6F}} \color[HTML]{000000} 0.923 & {\cellcolor[HTML]{9DD569}} \color[HTML]{000000} 792.903 & {\cellcolor[HTML]{3FAA59}} \color[HTML]{F1F1F1} 0.208 & {\cellcolor[HTML]{FEE08B}} \color[HTML]{000000} 1.800 & {\cellcolor[HTML]{D9EF8B}} \color[HTML]{000000} 10 & \bfseries normal \\
45 & {\cellcolor[HTML]{42AC5A}} \color[HTML]{F1F1F1} 0.925 & {\cellcolor[HTML]{279F53}} \color[HTML]{F1F1F1} 781.354 & \bfseries {\cellcolor[HTML]{006837}} \color[HTML]{F1F1F1} 0.206 & \bfseries {\cellcolor[HTML]{006837}} \color[HTML]{F1F1F1} 0.000 & {\cellcolor[HTML]{A50026}} \color[HTML]{F1F1F1} 16 & \bfseries normal \\
50 & {\cellcolor[HTML]{91D068}} \color[HTML]{000000} 0.923 & {\cellcolor[HTML]{7FC866}} \color[HTML]{000000} 789.607 & {\cellcolor[HTML]{73C264}} \color[HTML]{000000} 0.208 & \bfseries {\cellcolor[HTML]{006837}} \color[HTML]{F1F1F1} 0.000 & {\cellcolor[HTML]{66BD63}} \color[HTML]{F1F1F1} 8 & \bfseries normal \\
55 & {\cellcolor[HTML]{54B45F}} \color[HTML]{F1F1F1} 0.925 & {\cellcolor[HTML]{199750}} \color[HTML]{F1F1F1} 780.039 & {\cellcolor[HTML]{17934E}} \color[HTML]{F1F1F1} 0.207 & \bfseries {\cellcolor[HTML]{006837}} \color[HTML]{F1F1F1} 0.000 & {\cellcolor[HTML]{FEE08B}} \color[HTML]{000000} 12 & \bfseries normal \\
60 & {\cellcolor[HTML]{75C465}} \color[HTML]{000000} 0.924 & {\cellcolor[HTML]{42AC5A}} \color[HTML]{F1F1F1} 783.824 & {\cellcolor[HTML]{9BD469}} \color[HTML]{000000} 0.209 & {\cellcolor[HTML]{FEE08B}} \color[HTML]{000000} 1.800 & \bfseries {\cellcolor[HTML]{006837}} \color[HTML]{F1F1F1} 6 & \bfseries normal \\
65 & {\cellcolor[HTML]{BBE278}} \color[HTML]{000000} 0.922 & {\cellcolor[HTML]{98D368}} \color[HTML]{000000} 792.220 & {\cellcolor[HTML]{17934E}} \color[HTML]{F1F1F1} 0.207 & {\cellcolor[HTML]{FEE08B}} \color[HTML]{000000} 1.800 & {\cellcolor[HTML]{FEE08B}} \color[HTML]{000000} 12 & \bfseries normal \\
70 & \bfseries {\cellcolor[HTML]{006837}} \color[HTML]{F1F1F1} 0.927 & {\cellcolor[HTML]{8ECF67}} \color[HTML]{000000} 791.150 & {\cellcolor[HTML]{36A657}} \color[HTML]{F1F1F1} 0.208 & {\cellcolor[HTML]{A50026}} \color[HTML]{F1F1F1} 3.000 & {\cellcolor[HTML]{F46D43}} \color[HTML]{F1F1F1} 14 & \bfseries normal \\
75 & {\cellcolor[HTML]{D9EF8B}} \color[HTML]{000000} 0.921 & {\cellcolor[HTML]{CFEB85}} \color[HTML]{000000} 799.398 & {\cellcolor[HTML]{57B65F}} \color[HTML]{F1F1F1} 0.208 & {\cellcolor[HTML]{A50026}} \color[HTML]{F1F1F1} 3.000 & {\cellcolor[HTML]{D9EF8B}} \color[HTML]{000000} 10 & \bfseries normal \\
80 & {\cellcolor[HTML]{C9E881}} \color[HTML]{000000} 0.922 & {\cellcolor[HTML]{BDE379}} \color[HTML]{000000} 796.864 & {\cellcolor[HTML]{6EC064}} \color[HTML]{000000} 0.208 & {\cellcolor[HTML]{A50026}} \color[HTML]{F1F1F1} 3.000 & {\cellcolor[HTML]{66BD63}} \color[HTML]{F1F1F1} 8 & \bfseries normal \\
85 & {\cellcolor[HTML]{A7D96B}} \color[HTML]{000000} 0.923 & {\cellcolor[HTML]{96D268}} \color[HTML]{000000} 792.106 & {\cellcolor[HTML]{73C264}} \color[HTML]{000000} 0.208 & {\cellcolor[HTML]{FEE08B}} \color[HTML]{000000} 1.800 & {\cellcolor[HTML]{66BD63}} \color[HTML]{F1F1F1} 8 & \bfseries normal \\
90 & {\cellcolor[HTML]{66BD63}} \color[HTML]{F1F1F1} 0.924 & {\cellcolor[HTML]{4EB15D}} \color[HTML]{F1F1F1} 784.736 & {\cellcolor[HTML]{0A7B41}} \color[HTML]{F1F1F1} 0.207 & {\cellcolor[HTML]{FFFEBE}} \color[HTML]{000000} 1.500 & {\cellcolor[HTML]{A50026}} \color[HTML]{F1F1F1} 16 & \bfseries normal \\
95 & {\cellcolor[HTML]{219C52}} \color[HTML]{F1F1F1} 0.925 & {\cellcolor[HTML]{118848}} \color[HTML]{F1F1F1} 778.020 & {\cellcolor[HTML]{036E3A}} \color[HTML]{F1F1F1} 0.206 & {\cellcolor[HTML]{D9EF8B}} \color[HTML]{000000} 1.200 & {\cellcolor[HTML]{A50026}} \color[HTML]{F1F1F1} 16 & \bfseries normal \\
100 & {\cellcolor[HTML]{A0D669}} \color[HTML]{000000} 0.923 & {\cellcolor[HTML]{FEE593}} \color[HTML]{000000} 813.366 & {\cellcolor[HTML]{7AC665}} \color[HTML]{000000} 0.209 & {\cellcolor[HTML]{D9EF8B}} \color[HTML]{000000} 1.200 & {\cellcolor[HTML]{F46D43}} \color[HTML]{F1F1F1} 14 & \bfseries normal \\
105 & {\cellcolor[HTML]{2AA054}} \color[HTML]{F1F1F1} 0.925 & {\cellcolor[HTML]{D9EF8B}} \color[HTML]{000000} 800.746 & {\cellcolor[HTML]{73C264}} \color[HTML]{000000} 0.208 & {\cellcolor[HTML]{FFFEBE}} \color[HTML]{000000} 1.500 & {\cellcolor[HTML]{F46D43}} \color[HTML]{F1F1F1} 14 & \bfseries normal \\
110 & {\cellcolor[HTML]{87CB67}} \color[HTML]{000000} 0.923 & {\cellcolor[HTML]{73C264}} \color[HTML]{000000} 788.276 & {\cellcolor[HTML]{75C465}} \color[HTML]{000000} 0.208 & {\cellcolor[HTML]{FEE08B}} \color[HTML]{000000} 1.800 & {\cellcolor[HTML]{66BD63}} \color[HTML]{F1F1F1} 8 & \bfseries normal \\
115 & {\cellcolor[HTML]{E5F49B}} \color[HTML]{000000} 0.921 & {\cellcolor[HTML]{DDF191}} \color[HTML]{000000} 801.401 & {\cellcolor[HTML]{7AC665}} \color[HTML]{000000} 0.209 & {\cellcolor[HTML]{FFFEBE}} \color[HTML]{000000} 1.500 & {\cellcolor[HTML]{D9EF8B}} \color[HTML]{000000} 10 & \bfseries normal \\
120 & {\cellcolor[HTML]{E3F399}} \color[HTML]{000000} 0.921 & {\cellcolor[HTML]{DCF08F}} \color[HTML]{000000} 801.258 & {\cellcolor[HTML]{75C465}} \color[HTML]{000000} 0.208 & {\cellcolor[HTML]{D9EF8B}} \color[HTML]{000000} 1.200 & {\cellcolor[HTML]{D9EF8B}} \color[HTML]{000000} 10 & \bfseries normal \\
125 & {\cellcolor[HTML]{BBE278}} \color[HTML]{000000} 0.922 & {\cellcolor[HTML]{96D268}} \color[HTML]{000000} 792.040 & {\cellcolor[HTML]{07753E}} \color[HTML]{F1F1F1} 0.207 & {\cellcolor[HTML]{D9EF8B}} \color[HTML]{000000} 1.200 & {\cellcolor[HTML]{FEE08B}} \color[HTML]{000000} 12 & \bfseries normal \\
130 & {\cellcolor[HTML]{D1EC86}} \color[HTML]{000000} 0.922 & {\cellcolor[HTML]{B1DE71}} \color[HTML]{000000} 795.227 & {\cellcolor[HTML]{18954F}} \color[HTML]{F1F1F1} 0.207 & {\cellcolor[HTML]{FFFEBE}} \color[HTML]{000000} 1.500 & {\cellcolor[HTML]{FEE08B}} \color[HTML]{000000} 12 & \bfseries normal \\
135 & {\cellcolor[HTML]{60BA62}} \color[HTML]{F1F1F1} 0.924 & {\cellcolor[HTML]{279F53}} \color[HTML]{F1F1F1} 781.505 & {\cellcolor[HTML]{E6F59D}} \color[HTML]{000000} 0.210 & {\cellcolor[HTML]{D9EF8B}} \color[HTML]{000000} 1.200 & \bfseries {\cellcolor[HTML]{006837}} \color[HTML]{F1F1F1} 6 & \bfseries normal \\
140 & {\cellcolor[HTML]{A2D76A}} \color[HTML]{000000} 0.923 & {\cellcolor[HTML]{78C565}} \color[HTML]{000000} 788.882 & {\cellcolor[HTML]{CDEA83}} \color[HTML]{000000} 0.210 & {\cellcolor[HTML]{FFFEBE}} \color[HTML]{000000} 1.500 & \bfseries {\cellcolor[HTML]{006837}} \color[HTML]{F1F1F1} 6 & \bfseries normal \\
228 & {\cellcolor[HTML]{A50026}} \color[HTML]{F1F1F1} 0.913 & {\cellcolor[HTML]{A50026}} \color[HTML]{F1F1F1} 841.743 & {\cellcolor[HTML]{A50026}} \color[HTML]{F1F1F1} 0.216 & \bfseries {\cellcolor[HTML]{006837}} \color[HTML]{F1F1F1} 0.000 & {\cellcolor[HTML]{66BD63}} \color[HTML]{F1F1F1} 8 & random \\
230 & {\cellcolor[HTML]{E44C34}} \color[HTML]{F1F1F1} 0.915 & {\cellcolor[HTML]{F16640}} \color[HTML]{F1F1F1} 828.734 & {\cellcolor[HTML]{B91326}} \color[HTML]{F1F1F1} 0.215 & \bfseries {\cellcolor[HTML]{006837}} \color[HTML]{F1F1F1} 0.000 & {\cellcolor[HTML]{FEE08B}} \color[HTML]{000000} 12 & random \\
232 & {\cellcolor[HTML]{DD3D2D}} \color[HTML]{F1F1F1} 0.915 & {\cellcolor[HTML]{DD3D2D}} \color[HTML]{F1F1F1} 833.265 & {\cellcolor[HTML]{AB0626}} \color[HTML]{F1F1F1} 0.216 & \bfseries {\cellcolor[HTML]{006837}} \color[HTML]{F1F1F1} 0.000 & {\cellcolor[HTML]{A50026}} \color[HTML]{F1F1F1} 16 & random \\
\bottomrule
\end{tabular}
\end{table}

\begin{table}[ht!]
\centering
\caption{Experiment using PD-TGCN}
\label{tab:PDTCExperiments}
\begin{tabular}{lrrrrrl}
\toprule
 & R2 & RMSE & MAPE & weight & k & knhs \\
\midrule
1 & {\cellcolor[HTML]{FFFEBE}} \color[HTML]{000000} 0.922 & {\cellcolor[HTML]{FBFDBA}} \color[HTML]{000000} 793.945 & {\cellcolor[HTML]{FEEB9D}} \color[HTML]{000000} 0.208 & {\cellcolor[HTML]{A50026}} \color[HTML]{F1F1F1} 3.000 & \bfseries {\cellcolor[HTML]{006837}} \color[HTML]{F1F1F1} 6 & normal \\
6 & \bfseries {\cellcolor[HTML]{006837}} \color[HTML]{F1F1F1} 0.928 & \bfseries {\cellcolor[HTML]{006837}} \color[HTML]{F1F1F1} 766.888 & \bfseries {\cellcolor[HTML]{006837}} \color[HTML]{F1F1F1} 0.204 & {\cellcolor[HTML]{A50026}} \color[HTML]{F1F1F1} 3.000 & {\cellcolor[HTML]{A50026}} \color[HTML]{F1F1F1} 16 & normal \\
11 & {\cellcolor[HTML]{FCA55D}} \color[HTML]{000000} 0.919 & {\cellcolor[HTML]{F8864F}} \color[HTML]{F1F1F1} 808.776 & {\cellcolor[HTML]{B9E176}} \color[HTML]{000000} 0.206 & {\cellcolor[HTML]{FFFEBE}} \color[HTML]{000000} 1.500 & {\cellcolor[HTML]{66BD63}} \color[HTML]{F1F1F1} 8 & normal \\
16 & {\cellcolor[HTML]{FFF0A6}} \color[HTML]{000000} 0.921 & {\cellcolor[HTML]{FEE491}} \color[HTML]{000000} 799.328 & {\cellcolor[HTML]{B5DF74}} \color[HTML]{000000} 0.206 & {\cellcolor[HTML]{D9EF8B}} \color[HTML]{000000} 1.200 & {\cellcolor[HTML]{66BD63}} \color[HTML]{F1F1F1} 8 & normal \\
21 & {\cellcolor[HTML]{A5D86A}} \color[HTML]{000000} 0.924 & {\cellcolor[HTML]{93D168}} \color[HTML]{000000} 781.797 & {\cellcolor[HTML]{42AC5A}} \color[HTML]{F1F1F1} 0.205 & {\cellcolor[HTML]{A50026}} \color[HTML]{F1F1F1} 3.000 & {\cellcolor[HTML]{FEE08B}} \color[HTML]{000000} 12 & normal \\
26 & {\cellcolor[HTML]{FEC877}} \color[HTML]{000000} 0.920 & {\cellcolor[HTML]{FDB365}} \color[HTML]{000000} 804.999 & {\cellcolor[HTML]{E9F6A1}} \color[HTML]{000000} 0.207 & {\cellcolor[HTML]{A50026}} \color[HTML]{F1F1F1} 3.000 & {\cellcolor[HTML]{66BD63}} \color[HTML]{F1F1F1} 8 & normal \\
31 & {\cellcolor[HTML]{2DA155}} \color[HTML]{F1F1F1} 0.926 & {\cellcolor[HTML]{36A657}} \color[HTML]{F1F1F1} 774.598 & {\cellcolor[HTML]{17934E}} \color[HTML]{F1F1F1} 0.205 & {\cellcolor[HTML]{FEE08B}} \color[HTML]{000000} 1.800 & {\cellcolor[HTML]{A50026}} \color[HTML]{F1F1F1} 16 & normal \\
36 & {\cellcolor[HTML]{E5F49B}} \color[HTML]{000000} 0.923 & {\cellcolor[HTML]{E14430}} \color[HTML]{F1F1F1} 814.657 & {\cellcolor[HTML]{BBE278}} \color[HTML]{000000} 0.206 & {\cellcolor[HTML]{FEE08B}} \color[HTML]{000000} 1.800 & {\cellcolor[HTML]{F46D43}} \color[HTML]{F1F1F1} 14 & normal \\
41 & {\cellcolor[HTML]{FDB96A}} \color[HTML]{000000} 0.920 & {\cellcolor[HTML]{FB9D59}} \color[HTML]{000000} 806.809 & {\cellcolor[HTML]{B1DE71}} \color[HTML]{000000} 0.206 & {\cellcolor[HTML]{FEE08B}} \color[HTML]{000000} 1.800 & {\cellcolor[HTML]{D9EF8B}} \color[HTML]{000000} 10 & normal \\
46 & {\cellcolor[HTML]{8CCD67}} \color[HTML]{000000} 0.925 & {\cellcolor[HTML]{98D368}} \color[HTML]{000000} 782.398 & {\cellcolor[HTML]{93D168}} \color[HTML]{000000} 0.206 & \bfseries {\cellcolor[HTML]{006837}} \color[HTML]{F1F1F1} 0.000 & {\cellcolor[HTML]{A50026}} \color[HTML]{F1F1F1} 16 & normal \\
51 & {\cellcolor[HTML]{FEDE89}} \color[HTML]{000000} 0.921 & {\cellcolor[HTML]{FECA79}} \color[HTML]{000000} 802.393 & {\cellcolor[HTML]{B9E176}} \color[HTML]{000000} 0.206 & \bfseries {\cellcolor[HTML]{006837}} \color[HTML]{F1F1F1} 0.000 & {\cellcolor[HTML]{66BD63}} \color[HTML]{F1F1F1} 8 & normal \\
56 & {\cellcolor[HTML]{EEF8A8}} \color[HTML]{000000} 0.922 & {\cellcolor[HTML]{E6F59D}} \color[HTML]{000000} 790.932 & {\cellcolor[HTML]{7DC765}} \color[HTML]{000000} 0.206 & \bfseries {\cellcolor[HTML]{006837}} \color[HTML]{F1F1F1} 0.000 & {\cellcolor[HTML]{FEE08B}} \color[HTML]{000000} 12 & normal \\
61 & {\cellcolor[HTML]{BDE379}} \color[HTML]{000000} 0.924 & {\cellcolor[HTML]{B3DF72}} \color[HTML]{000000} 784.820 & {\cellcolor[HTML]{FEDA86}} \color[HTML]{000000} 0.208 & {\cellcolor[HTML]{FEE08B}} \color[HTML]{000000} 1.800 & \bfseries {\cellcolor[HTML]{006837}} \color[HTML]{F1F1F1} 6 & normal \\
66 & {\cellcolor[HTML]{E8F59F}} \color[HTML]{000000} 0.923 & {\cellcolor[HTML]{E0F295}} \color[HTML]{000000} 790.054 & {\cellcolor[HTML]{108647}} \color[HTML]{F1F1F1} 0.204 & {\cellcolor[HTML]{FEE08B}} \color[HTML]{000000} 1.800 & {\cellcolor[HTML]{FEE08B}} \color[HTML]{000000} 12 & normal \\
71 & {\cellcolor[HTML]{E0F295}} \color[HTML]{000000} 0.923 & {\cellcolor[HTML]{E54E35}} \color[HTML]{F1F1F1} 813.824 & {\cellcolor[HTML]{BBE278}} \color[HTML]{000000} 0.206 & {\cellcolor[HTML]{A50026}} \color[HTML]{F1F1F1} 3.000 & {\cellcolor[HTML]{F46D43}} \color[HTML]{F1F1F1} 14 & normal \\
76 & {\cellcolor[HTML]{F88C51}} \color[HTML]{F1F1F1} 0.919 & {\cellcolor[HTML]{F46D43}} \color[HTML]{F1F1F1} 810.996 & {\cellcolor[HTML]{98D368}} \color[HTML]{000000} 0.206 & {\cellcolor[HTML]{A50026}} \color[HTML]{F1F1F1} 3.000 & {\cellcolor[HTML]{D9EF8B}} \color[HTML]{000000} 10 & normal \\
81 & {\cellcolor[HTML]{FED27F}} \color[HTML]{000000} 0.920 & {\cellcolor[HTML]{FDBF6F}} \color[HTML]{000000} 803.723 & {\cellcolor[HTML]{E9F6A1}} \color[HTML]{000000} 0.207 & {\cellcolor[HTML]{A50026}} \color[HTML]{F1F1F1} 3.000 & {\cellcolor[HTML]{66BD63}} \color[HTML]{F1F1F1} 8 & normal \\
86 & {\cellcolor[HTML]{FEDE89}} \color[HTML]{000000} 0.921 & {\cellcolor[HTML]{FECA79}} \color[HTML]{000000} 802.361 & {\cellcolor[HTML]{D9EF8B}} \color[HTML]{000000} 0.207 & {\cellcolor[HTML]{FEE08B}} \color[HTML]{000000} 1.800 & {\cellcolor[HTML]{66BD63}} \color[HTML]{F1F1F1} 8 & normal \\
91 & {\cellcolor[HTML]{75C465}} \color[HTML]{000000} 0.925 & {\cellcolor[HTML]{82C966}} \color[HTML]{000000} 780.344 & {\cellcolor[HTML]{57B65F}} \color[HTML]{F1F1F1} 0.205 & {\cellcolor[HTML]{FFFEBE}} \color[HTML]{000000} 1.500 & {\cellcolor[HTML]{A50026}} \color[HTML]{F1F1F1} 16 & normal \\
96 & {\cellcolor[HTML]{84CA66}} \color[HTML]{000000} 0.925 & {\cellcolor[HTML]{91D068}} \color[HTML]{000000} 781.653 & {\cellcolor[HTML]{138C4A}} \color[HTML]{F1F1F1} 0.204 & {\cellcolor[HTML]{D9EF8B}} \color[HTML]{000000} 1.200 & {\cellcolor[HTML]{A50026}} \color[HTML]{F1F1F1} 16 & normal \\
101 & {\cellcolor[HTML]{73C264}} \color[HTML]{000000} 0.925 & {\cellcolor[HTML]{FECE7C}} \color[HTML]{000000} 801.828 & {\cellcolor[HTML]{A0D669}} \color[HTML]{000000} 0.206 & {\cellcolor[HTML]{D9EF8B}} \color[HTML]{000000} 1.200 & {\cellcolor[HTML]{F46D43}} \color[HTML]{F1F1F1} 14 & normal \\
106 & {\cellcolor[HTML]{9BD469}} \color[HTML]{000000} 0.924 & {\cellcolor[HTML]{FCAA5F}} \color[HTML]{000000} 805.740 & {\cellcolor[HTML]{A0D669}} \color[HTML]{000000} 0.206 & {\cellcolor[HTML]{FFFEBE}} \color[HTML]{000000} 1.500 & {\cellcolor[HTML]{F46D43}} \color[HTML]{F1F1F1} 14 & normal \\
111 & {\cellcolor[HTML]{FED481}} \color[HTML]{000000} 0.920 & {\cellcolor[HTML]{FDBF6F}} \color[HTML]{000000} 803.613 & {\cellcolor[HTML]{E8F59F}} \color[HTML]{000000} 0.207 & {\cellcolor[HTML]{FEE08B}} \color[HTML]{000000} 1.800 & {\cellcolor[HTML]{66BD63}} \color[HTML]{F1F1F1} 8 & normal \\
116 & {\cellcolor[HTML]{FECA79}} \color[HTML]{000000} 0.920 & {\cellcolor[HTML]{FDB567}} \color[HTML]{000000} 804.636 & {\cellcolor[HTML]{BDE379}} \color[HTML]{000000} 0.206 & {\cellcolor[HTML]{FFFEBE}} \color[HTML]{000000} 1.500 & {\cellcolor[HTML]{D9EF8B}} \color[HTML]{000000} 10 & normal \\
121 & {\cellcolor[HTML]{FEE797}} \color[HTML]{000000} 0.921 & {\cellcolor[HTML]{FED683}} \color[HTML]{000000} 800.967 & {\cellcolor[HTML]{B1DE71}} \color[HTML]{000000} 0.206 & {\cellcolor[HTML]{D9EF8B}} \color[HTML]{000000} 1.200 & {\cellcolor[HTML]{D9EF8B}} \color[HTML]{000000} 10 & normal \\
126 & {\cellcolor[HTML]{BDE379}} \color[HTML]{000000} 0.924 & {\cellcolor[HTML]{B1DE71}} \color[HTML]{000000} 784.622 & {\cellcolor[HTML]{36A657}} \color[HTML]{F1F1F1} 0.205 & {\cellcolor[HTML]{D9EF8B}} \color[HTML]{000000} 1.200 & {\cellcolor[HTML]{FEE08B}} \color[HTML]{000000} 12 & normal \\
131 & {\cellcolor[HTML]{69BE63}} \color[HTML]{F1F1F1} 0.925 & {\cellcolor[HTML]{4EB15D}} \color[HTML]{F1F1F1} 776.349 & {\cellcolor[HTML]{3CA959}} \color[HTML]{F1F1F1} 0.205 & {\cellcolor[HTML]{FFFEBE}} \color[HTML]{000000} 1.500 & {\cellcolor[HTML]{FEE08B}} \color[HTML]{000000} 12 & normal \\
136 & {\cellcolor[HTML]{D5ED88}} \color[HTML]{000000} 0.923 & {\cellcolor[HTML]{CBE982}} \color[HTML]{000000} 787.485 & {\cellcolor[HTML]{FFF1A8}} \color[HTML]{000000} 0.208 & {\cellcolor[HTML]{D9EF8B}} \color[HTML]{000000} 1.200 & \bfseries {\cellcolor[HTML]{006837}} \color[HTML]{F1F1F1} 6 & normal \\
141 & {\cellcolor[HTML]{E3F399}} \color[HTML]{000000} 0.923 & {\cellcolor[HTML]{DDF191}} \color[HTML]{000000} 789.619 & {\cellcolor[HTML]{FFF1A8}} \color[HTML]{000000} 0.208 & {\cellcolor[HTML]{FFFEBE}} \color[HTML]{000000} 1.500 & \bfseries {\cellcolor[HTML]{006837}} \color[HTML]{F1F1F1} 6 & normal \\
225 & {\cellcolor[HTML]{FEEC9F}} \color[HTML]{000000} 0.921 & {\cellcolor[HTML]{FEE08B}} \color[HTML]{000000} 800.011 & {\cellcolor[HTML]{F67C4A}} \color[HTML]{F1F1F1} 0.210 & \bfseries {\cellcolor[HTML]{006837}} \color[HTML]{F1F1F1} 0.000 & {\cellcolor[HTML]{66BD63}} \color[HTML]{F1F1F1} 8 & \bfseries geo \\
226 & {\cellcolor[HTML]{A50026}} \color[HTML]{F1F1F1} 0.916 & {\cellcolor[HTML]{A50026}} \color[HTML]{F1F1F1} 822.098 & {\cellcolor[HTML]{CA2427}} \color[HTML]{F1F1F1} 0.211 & \bfseries {\cellcolor[HTML]{006837}} \color[HTML]{F1F1F1} 0.000 & {\cellcolor[HTML]{FEE08B}} \color[HTML]{000000} 12 & \bfseries geo \\
227 & {\cellcolor[HTML]{FA9656}} \color[HTML]{000000} 0.919 & {\cellcolor[HTML]{F57547}} \color[HTML]{F1F1F1} 810.243 & {\cellcolor[HTML]{A50026}} \color[HTML]{F1F1F1} 0.211 & \bfseries {\cellcolor[HTML]{006837}} \color[HTML]{F1F1F1} 0.000 & {\cellcolor[HTML]{A50026}} \color[HTML]{F1F1F1} 16 & \bfseries geo \\
229 & {\cellcolor[HTML]{FDBB6C}} \color[HTML]{000000} 0.920 & {\cellcolor[HTML]{FBA35C}} \color[HTML]{000000} 806.569 & {\cellcolor[HTML]{EE613E}} \color[HTML]{F1F1F1} 0.210 & \bfseries {\cellcolor[HTML]{006837}} \color[HTML]{F1F1F1} 0.000 & {\cellcolor[HTML]{66BD63}} \color[HTML]{F1F1F1} 8 & random \\
231 & {\cellcolor[HTML]{B50F26}} \color[HTML]{F1F1F1} 0.917 & {\cellcolor[HTML]{B50F26}} \color[HTML]{F1F1F1} 820.268 & {\cellcolor[HTML]{CA2427}} \color[HTML]{F1F1F1} 0.211 & \bfseries {\cellcolor[HTML]{006837}} \color[HTML]{F1F1F1} 0.000 & {\cellcolor[HTML]{FEE08B}} \color[HTML]{000000} 12 & random \\
233 & {\cellcolor[HTML]{E95538}} \color[HTML]{F1F1F1} 0.918 & {\cellcolor[HTML]{D93429}} \color[HTML]{F1F1F1} 816.082 & {\cellcolor[HTML]{A50026}} \color[HTML]{F1F1F1} 0.211 & \bfseries {\cellcolor[HTML]{006837}} \color[HTML]{F1F1F1} 0.000 & {\cellcolor[HTML]{A50026}} \color[HTML]{F1F1F1} 16 & random \\
\bottomrule
\end{tabular}
\end{table}

\end{document}